\documentclass[conference]{IEEEtran}
\usepackage{times}

\usepackage[numbers]{natbib}
\usepackage{multicol}
\usepackage[bookmarks=true]{hyperref}

\usepackage{url} 
\usepackage{amssymb}
\usepackage{amsmath}
\usepackage{amssymb}
\usepackage{multirow}
\usepackage{graphicx}       
\usepackage{booktabs}       
\usepackage{amsfonts}       
\usepackage[margin=1in]{geometry} 
\usepackage{booktabs}
\usepackage{adjustbox}
\usepackage{subcaption}
\usepackage{amssymb}
\usepackage{float}
\usepackage{cleveref}
\usepackage[table]{xcolor}
\usepackage{graphicx}
\usepackage{xcolor}
\usepackage{wasysym}
\usepackage{cuted} 
\usepackage{caption} 

\begin{document}

\title{Unified Embodied VLM Reasoning with Robotic Action via Autoregressive Discretized Pre-training}

\author{Yi Liu$^{1*}$, Sukai Wang$^{1*\dagger}$, Dafeng Wei$^{1}$, Xiaowei Cai$^{1}$, Linqing Zhong$^{1}$, Jiange Yang$^{1}$, Guanghui Ren$^{2}$, \\[0.2em]
Jinyu Zhang$^{1,3}$, Maoqing Yao$^{2}$, Chuankang Li$^{1}$, Xindong He$^{1}$, Liliang Chen$^{1}$, Jianlan Luo$^{1,3\ddagger}$ \\[0.5em]
$^{1}$AgiBot Research \qquad $^{2}$AgiBot \qquad $^{3}$Shanghai Innovation Institute \\
\thanks{$^{*}$Equal contribution \quad $^{\dagger}$Project leader \quad $^{\ddagger}$Corresponding author}
}



%

\maketitle
\begin{strip}
    \centering
    \includegraphics[width=0.9\linewidth]{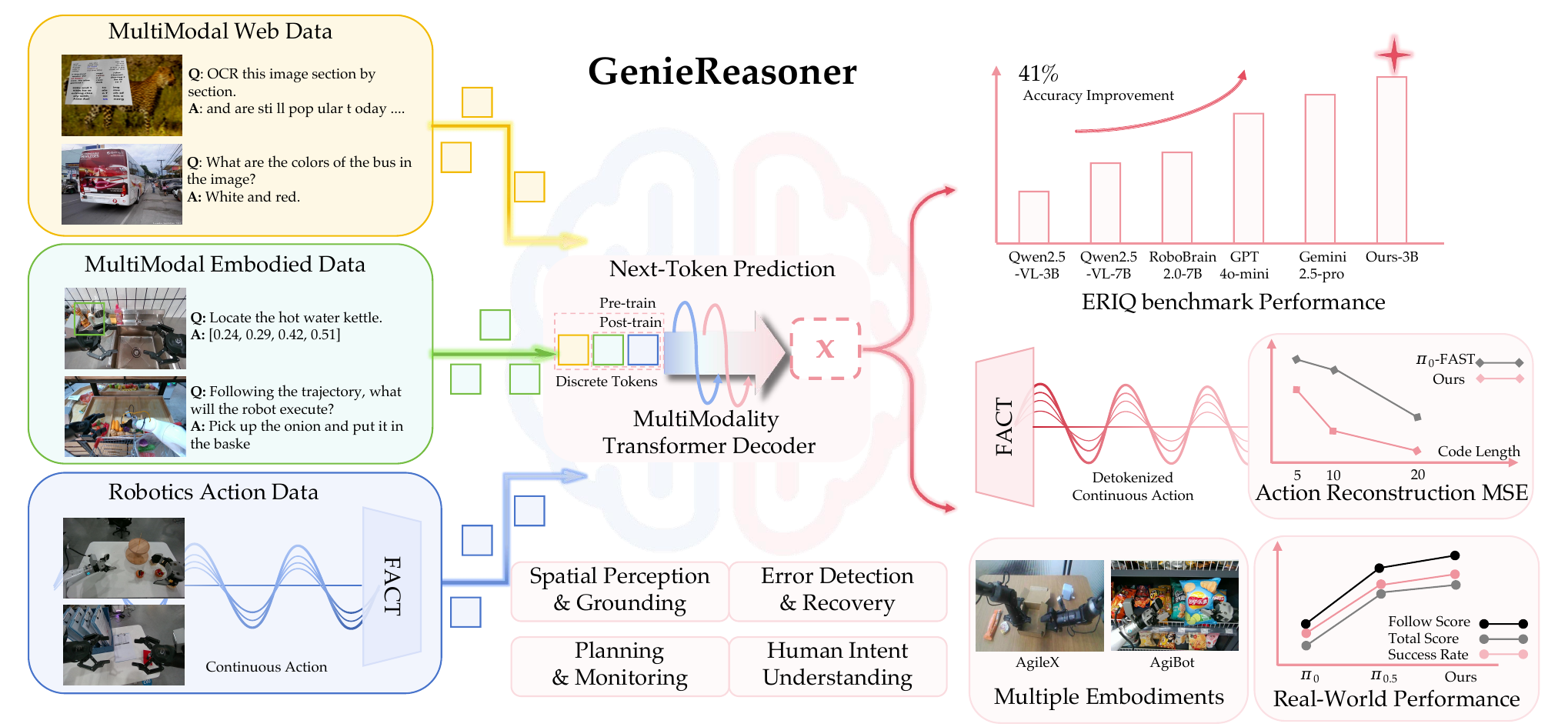}
    \captionof{figure}{Overview of GenieReasoner. (Left) We leverage large-scale general and embodied multimodal data alongside FACT, a flow-matching tokenizer that discretizes continuous actions for alignment with semantic inputs. (Center) A unified autoregressive transformer co-optimizes high-level reasoning and low-level control via next-token prediction. (Right) This design yields state-of-the-art results: a 41\% accuracy improvement on ERIQ, superior reconstruction fidelity compared to $\pi_0$-FAST, and robust real-world generalization outperforming flow-based baselines like $\pi_{0.5}$.}
    \label{fig:fig1}
\end{strip}

\setcounter{figure}{1}

\begin{abstract}
General-purpose robotic systems operating in open-world environments must achieve both broad generalization and high-precision action execution, a combination that remains challenging for existing Vision-Language-Action (VLA) models. While large Vision-Language Models (VLMs) improve semantic generalization, insufficient embodied reasoning leads to brittle behavior, and conversely, strong reasoning alone is inadequate without precise control.
To provide a decoupled and quantitative assessment of this bottleneck, we introduce Embodied Reasoning Intelligence Quotient (ERIQ), a large-scale embodied reasoning benchmark in robotic manipulation, comprising 6K+ question–answer pairs across four reasoning dimensions.
By decoupling reasoning from execution, ERIQ enables systematic evaluation and reveals a strong positive correlation between embodied reasoning capability and end-to-end VLA generalization.
To bridge the gap from reasoning to precise execution, we propose FACT, a flow-matching-based action tokenizer that converts continuous control into discrete sequences while preserving high-fidelity trajectory reconstruction. 
The resulting GenieReasoner jointly optimizes reasoning and action in a unified space, outperforming both continuous-action and prior discrete-action baselines in real-world tasks.
Together, ERIQ and FACT provide a principled framework for diagnosing and overcoming the reasoning–precision trade-off, advancing robust, general-purpose robotic manipulation. 
\end{abstract}

\IEEEpeerreviewmaketitle

\section{Introduction}

The pursuit of general-purpose robotic systems capable of operating autonomously in unstructured, open-world environments represents a fundamental challenge in artificial intelligence~\cite{rt1, rt2, openvla, black2410pi0, bu2025agibot}.
This endeavor necessitates simultaneous achievement of two critical, yet often competing objectives: achieving \textit{broad generalization} across diverse, unseen scenarios~\cite{black2025pi05,gemini_robotics} while maintaining \textit{high-precision action execution} for reliable task completion~\cite{Octo2024, chi2025diffusion, zhao2023learning}.
Recent advances in Large Vision-Language Models (VLMs), pre-trained on extensive internet-scale datasets, have established a new research direction by enabling the integration of rich semantic knowledge derived from diverse multimodal data, thereby significantly enhancing generalization capabilities for downstream tasks.
These advances have led to Vision-Language-Action (VLA) models that extend VLMs' reasoning capability to directly predict robotic actions, unifying perception, reasoning, and control, establishing a principled bridge between abstract semantic reasoning and concrete low-level physical execution~\cite{black2410pi0, black2025pi05, bu2025agibot, Lin2025onetwovla, openvla, pertsch2025fast, rt1, rt2, lv2025f1, gemini_robotics, Zhou2025chatvla, qu2025spatialvla, zhao2024vlmpcvisionlanguagemodelpredictive, kim2025finetuning, liu2025hybridvla, shukor2025smolvla, zheng2024tracevla}.

We posit that robust \textit{embodied reasoning} is an indispensable prerequisite for achieving such generalization. This capability encompasses the synthesis of spatial awareness, temporal dynamics, and causal logic to explicitly plan, monitor, and adaptively correct physical actions. Without sufficient reasoning, learned policies are prone to falling back on superficial visual-action correlations. This leads to brittle performance under distribution shifts, novel object configurations, or unexpected environmental perturbations~\cite{zawalski2024robotic, Lin2025onetwovla, driess2025knowledge}. 
However, possessing strong reasoning is fundamentally insufficient in isolation. 
Real-world success equally demands \textit{high-precision action execution}, which we define as the ability to generate fine-grained control signals that faithfully translate reasoning into movement. 
Empirical observations suggest a persistent tension in current systems where models optimized for reasoning often exhibit reduced precision, while those achieving high-fidelity execution demonstrate limited generalization~\cite{driess2025knowledge, pertsch2025fast}.

To address the lack of a systematic way to diagnose this tension, we first introduce the Embodied Reasoning Intelligence Quotient (ERIQ), which stands as the first large-scale closed-loop benchmark for embodied reasoning in robotic manipulation.
ERIQ provides a solution to the evaluation bottleneck by decoupling cognitive reasoning from motor control, which enables the independent quantification of reasoning without confounding action execution errors.
The benchmark comprises over 6,000 question-answer pairs across four key dimensions: Spatial Perception \& Grounding, Task Planning \& Monitoring, Error Detection \& Recovery, and Human Intent Understanding. 
By incorporating reactive dimensions such as error recovery and intent anticipation, ERIQ provides a comprehensive lens into the ability of a model to reason within a dynamic, closed-loop environment.
Our empirical analysis on ERIQ reveals a strong positive correlation between VLM reasoning capabilities and end-to-end VLA performance, providing evidence that reasoning capacity is a primary driver of generalization.

The empirical insights gained from ERIQ validate that embodied reasoning is the primary driver of generalization, yet they also highlight a critical implementation gap: current architectures struggle to translate this reasoning into high-precision physical movement. 
Achieving such precision requires bridging discrete semantic representations from language-centric reasoning tokens with the continuous control signals necessary for physical execution. 
One common class of methods discretizes robotic actions to enable co-training with language tokens. However, this approach faces a steep precision-efficiency trade-off. Simple uniform binning necessitates an excessive number of tokens to achieve fine-grained control, which consumes valuable context space~\cite{rt2, openvla}. Learned quantization, such as VQ-VAE, offers compact codes but lacks the high-precision control required for dexterity, often failing to reproduce exact continuous signals~\cite{vqvla, szot2024grounding}. 
Furthermore, rule-based adaptive schemes like FAST rely on variable-length encodings, where the non-deterministic structure frequently leads to decoding failures~\cite{pertsch2025fast, stablefast}.
Alternative hybrid architectures, such as $\pi_0$, append continuous prediction heads to discrete model backbones~\cite{black2410pi0}. 
However, joint training with conflicting objectives often degrades performance~\cite{driess2025knowledge}.
The competition between discrete cross-entropy for semantic tokens and continuous regression for actions creates a training misalignment that can undermine the model's reasoning capabilities~\cite{driess2025knowledge}. 

To bridge the gap between reasoning and execution, we present the Flow-matching Action Tokenizer (FACT), a discrete action tokenizer that leverages flow-matching to reconstruct high-fidelity continuous trajectories from compact token sequences.
FACT transforms motor control into a discrete sequence modeling while preserving continuous-space precision. 
This enables the GenieReasoner(illustrated in \Cref{fig:fig1}) to co-optimize reasoning and action within a unified gradient space.
This alignment directly projects the VLM's reasoning capabilities into precise actions, avoiding the generalization penalty of continuous heads while surpassing the fidelity limits of existing discrete tokenizers.
Empirical evaluation demonstrates that GenieReasoner achieves superior performance 
over both continuous-action baselines (e.g., $\pi_{0.5}$~\cite{black2025pi05}) 
and discrete-action baselines (e.g., $\pi_0$-FAST~\cite{pertsch2025fast}), 
effectively balancing embodied reasoning and action precision across diverse 
benchmarks and real-world robotic deployments.

Our key contributions are twofold. First, we introduce ERIQ, the first large-scale closed-loop benchmark that provides empirical evidence that robust reasoning is essential for open-world generalization.
Second, we present FACT, a flow-matching action tokenizer that bridges discrete semantic representations with continuous control. 
The resulting GenieReasoner achieves state-of-the-art performance through unified co-optimization of embodied reasoning and precise action execution within a single gradient space.

\section{Related Works}
GenieReasoner bridges the gap between abstract semantic reasoning and high-precision physical control. We contextualize our approach by surveying the literature across two primary dimensions: embodied reasoning frameworks and the evolution of VLA architectures.

\subsection{Embodied Reasoning and Benchmarks}
As robotic systems scale beyond primitive execution, research has focused on the intersection of high-level reasoning and low-level control, necessitating robust embodied reasoning capabilities and benchmarks.

\paragraph{Hierarchical Planning} 
Early research in hierarchical planning utilized pre-trained or fine-tuned Large Language Models (LLMs) to decompose complex goals into high-level sub-tasks, which were then executed by specialized low-level policies~\cite{saycan, huang2022language, driess2023palm, huang2022inner, rth, liang2022code, zeng2022socratic, mees2022grounding, sharma2022skill, hirobot}. 
Subsequent efforts have focused on developing more structured reasoning systems.
Efforts like Gemini Robotics~\cite{gemini_robotics} and RoboBrain~\cite{robobrain} augment planning via auxiliary affordances or specialized fine-tuning.
However, separating high-level reasoning from low-level control often causes misalignment, where physical execution diverges from reasoning instructions.

\paragraph{Reasoning Integration and Co-training}
To bridge the gap between abstract planning and physical execution, recent frameworks integrate reasoning directly into the policy's inference pipeline, either through explicit intermediate traces or implicit large-scale co-training~\cite{black2025pi05, zawalski2024robotic, Lin2025onetwovla, Zhou2025chatvla, qu2025eo, ecot-light}. 
Embodied Chain-of-Thought (ECoT)~\cite{zawalski2024robotic} trains VLAs to generate explicit multi-step reasoning outputs before predicting actions. 
\(\pi_{0.5}\)~\cite{black2025pi05} employs a two-stage strategy that pre-trains on diverse multimodal data to acquire broad semantic priors before fine-tuning a continuous control expert. 
Recent works further explore reasoning and generalization through \(\pi_{0}\)-style co-training or purely autoregressive approaches~\cite{Lin2025onetwovla, Zhou2025chatvla, qu2025eo, yang2025instructvla}. 

\paragraph{Embodied Reasoning Benchmarks}
To measure embodied reasoning capabilities, the community has developed benchmarks spanning various levels of embodied understanding~\cite{sermanet2024robovqa, chen2023egoplan, li2024mmro, cheng2024egothink, yang2025embodiedbench, song2025robospatial}. 
Some benchmarks have focused on isolated embodied reasoning facets, such as egocentric perception~\cite{majumdar2024openeqa, cheng2024egothink}, long-horizon planning~\cite{sermanet2024robovqa, chen2023egoplan}, or affordance-aware placement~\cite{yuan2024robopoint, zhao2025manipbench}.
Recent works have moved toward comprehensive and systemic evaluation frameworks for embodied reasoning~\cite{gemini_robotics, robobrain, qu2025eo}. For instance, ERQA\cite{gemini_robotics} pioneered the integration of capabilities essential for physical world interaction, including spatial awareness, trajectory reasoning, and task-level awareness. 

\begin{figure*}[t]
    \centering
    \includegraphics[width=0.95\linewidth]{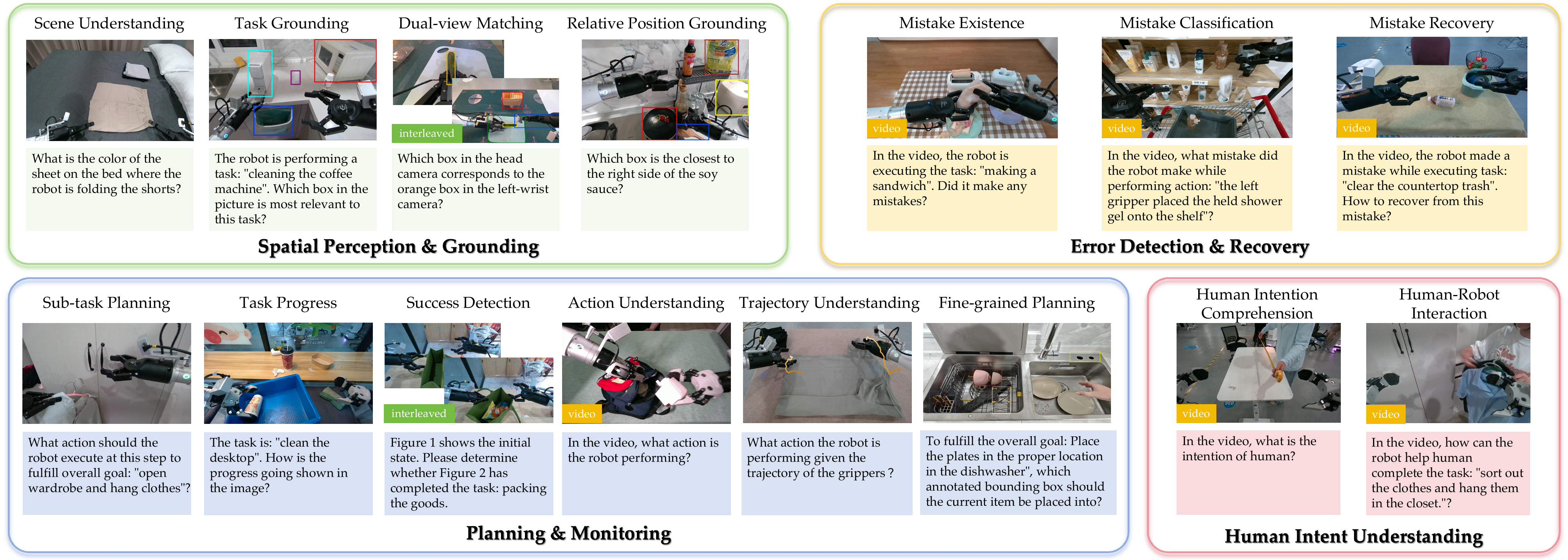}
    \caption{
    Illustration of the ERIQ benchmark. Example samples from the four major categories of embodied reasoning.
    }
    \label{fig:benchmark_samples}
    \vspace{-0.3cm} 
\end{figure*}

\subsection{Vision-Language-Action Models}
Current VLA architectures predominantly fall into two categories based on their action representation strategies.
\paragraph{Discrete Quantization and Autoregressive Tokenization}
A prominent class of VLAs transforms continuous robotic actions into discrete tokens, allowing the model to treat action prediction as a next-token prediction task similar to natural language \cite{rt2, openvla, zawalski2024robotic, vqvla, liang2025discrete, liu2025hybridvla, vla0, faster}.
While this approach benefits from the mature optimization landscape of Transformer-based VLMs, its maximum potential is inherently constrained by the action tokenizer. 
Concretely, simple uniform binning requires an excessively large vocabulary to achieve the resolution necessary for fine motor control~\cite{openvla}, whereas learned quantization methods suffer from imprecise execution~\cite{vqvla}. 
Recent attempts like FAST \cite{pertsch2025fast} utilize byte-pair encoding (BPE) to achieve compact action compression, which introduces variable-length sequence structures that destabilize autoregressive decoding~\cite{stablefast}.

\paragraph{Continuous Control via Generative Action Heads}
To achieve the high-fidelity control required for dexterous tasks, several recent works have integrated continuous action representations using diffusion or flow-matching objectives \cite{Octo2024, black2410pi0, bu2025agibot, bjorck2025gr00t, black2025pi05, shukor2025smolvla, shi2025memoryvla, qu2025eo, yang2025instructvla, liu2025hybridvla, chen2025internvlam1}. While these generative heads offer superior precision, they often face significant optimization conflicts when paired with discrete VLM backbones. Specifically, high-frequency gradients from the action-denoising objective can interfere with and degrade the VLM's pre-trained semantic reasoning capabilities. Current literature addresses this through complex architectural safeguards: Driess et al. \cite{driess2025knowledge} propose ``Knowledge Insulation'' to block gradient flow into the backbone, while ChatVLA \cite{Zhou2025chatvla} utilizes ``Phased Alignment'' to incrementally synchronize multimodal modalities. 

\section{Embodied Reasoning Intelligence Quotient Benchmark}
\label{vlm}
While broad generalization in robotics demands robust embodied reasoning, current evaluation methodologies suffer from a fundamental credit assignment problem. When a robotic system fails a task, it is often unclear whether the failure stemmed from a high-level cognitive misunderstanding or a low-level motor execution error. This ambiguity obscures the specific bottlenecks limiting model performance and hinders the systematic development of more capable systems. To resolve this, we introduce the Embodied Reasoning Intelligence Quotient (ERIQ) benchmark (\Cref{fig:benchmark_samples}), a diagnostic framework designed to isolate cognitive proficiency from physical dexterity. 
ERIQ comprises 6,052 Visual Question Answering (VQA) pairs specifically curated to quantify embodied comprehension independently of control policies. 
By evaluating a model's ability to process and reason about physical scenes without the confounding variables of continuous action execution, ERIQ provides a transparent metric for a model's cognitive readiness. The benchmark is organized into four core domains that encompass 15 fine-grained sub-tasks:

\paragraph{Spatial Perception and Grounding}
A prerequisite for deliberate robotic action is the ability to perceive and interpret the environment. 
This domain tests global and relational understanding through tasks such as \textit{Scene Understanding} and \textit{Relative Position Grounding}. These tasks assess an agent’s ability to resolve object referents from complex instructions (e.g., ``the bowl to the left of the cutting board'') and to solve the symbol grounding problem in cluttered scenes.

\paragraph{Task Planning and Execution Monitoring}
Intelligent agents must reason over causal structures to achieve complex goals. This domain includes \textit{Sub-task Planning} and \textit{Trajectory Analysis}, evaluating whether the agent can decompose high-level instructions into coherent action sequences and recognize task completion through visual state transitions.

\paragraph{Error Detection and Recovery}
Robust operation requires diagnosing and recovering from failures. We establish a hierarchy—\textit{Mistake Existence}, \textit{Classification}, and \textit{Recovery}, which measure the agent's ability to recognize errors, attribute causes, and propose corrective actions without requiring physical execution.

\paragraph{Human Intent Understanding}
Robots in human-centric environments must proactively anticipate needs rather than passively waiting for commands. 
We assess this by asking the agent to infer underlying goals and propose appropriate cooperative actions, treating social intelligence as a core reasoning challenge.

ERIQ comprises 6,052 VQA pairs derived entirely from authentic, first-person robotic trials across 100+ distinct task scenarios. More details are provided in Appendix~\ref{app:eriq_detail}.
Unlike existing benchmarks that offer fragmented support, ERIQ provides full coverage across four critical reasoning dimensions. Specifically, it addresses the "credit assignment problem" in robotics by incorporating \textit{Error Detection \& Recovery} and \textit{Human Intent Understanding} to evaluate closed-loop cognitive proficiency independently of motor errors. Detailed comparisons with other benchmarks are available in Appendix~\ref{app:eriq_compare}. As demonstrated in \Cref{sec:experiments}, our analysis reveals a strong positive correlation between ERIQ scores and end-to-end manipulation success, confirming that a robust embodied reasoning backbone is a primary driver of generalization in unstructured environments.

\begin{figure*}[t!]
    \centering
    \includegraphics[width=0.9\linewidth]{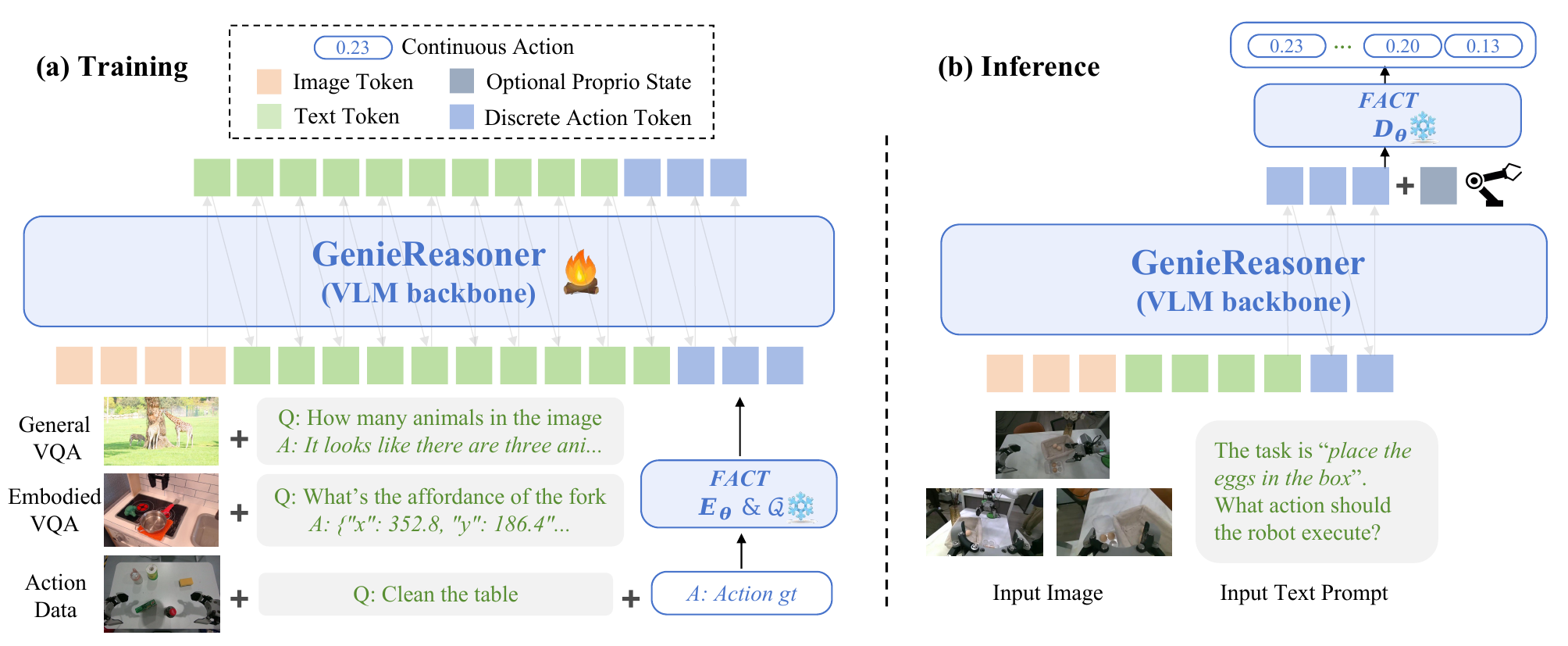}
    \caption{The GenieReasoner system architecture. (a) Training: Our unified pipeline jointly optimizes the VLM backbone for multimodal reasoning and robotic control by tokenizing continuous actions into a discrete latent space. This process encompasses the second and third stages of the training recipe (\Cref{sec:implementation}), where General VQA data is incorporated during the second stage to preserve foundational vision-language knowledge. 
    (b) Inference: Discrete action codes generated by the VLM backbone are decoded into continuous control signals via the FACT decoder, ensuring high-precision manipulation that is semantically grounded in the task instructions.}
    \vspace{-0.2in}
\label{fig:framework}
\end{figure*}

\section{GenieReasoner: Unified Discrete Low-level Policy Framework}
\subsection{Preliminaries}
We formulate the generalist robotic manipulation task as a learning problem for a policy $\pi_\theta$, parameterized by $\theta$. 
The system receives a multimodal observation stream and operates within a continuous action space $\mathcal{A} \subset \mathbb{R}^{S}$ (e.g., end-effector pose and gripper state). At each time step $\tau$, the observation $\mathbf{o}_{\tau}$ comprises $T_{obs}$ images $\mathbf{I}_{\tau} \in \mathbb{R}^{T_{obs} \times h \times w \times 3}$ and a natural language instruction $l$.

Following the Vision-Language-Action (VLA) route, we cast policy learning as a multimodal action generation task. 
Specifically, the objective is to maximize the log-likelihood of predicting future action chunks $\mathbf{a}_{\tau:\tau+H} \in \mathbb{R}^{H \times S}$ conditioned on the current observation and instruction, where $H$ denotes the prediction horizon.
The optimization objective over a demonstration dataset $\mathcal{D}$ is formulated as:
\begin{equation}
    \theta^* = \operatorname*{argmax}_{\theta} \mathbb{E}_{\zeta \sim \mathcal{D}} \left[ \sum_{\tau=1}^{|\zeta|} \log p_\theta(\mathbf{a}_{\tau:\tau+H} \mid \mathbf{I}_{\tau}, l) \right],
\end{equation}
where $|\zeta|$ denotes the episode length.
The model $\pi_\theta$ serves as a unified conditional probability estimator mapping high-dimensional inputs to low-level control.

\subsection{System Architecture}
To effectively ground the semantic reasoning of VLMs into high-precision robotic control, we must reconcile the conflict between the model's discrete autoregressive backbone and the continuous nature of motor execution. Existing discretization strategies face distinct limitations: uniform binning necessitates an excessive number of tokens to match the resolution required for manipulation~\cite{openvla, rt2}; learned quantization (e.g., VQ-VAE) achieves compactness but often lacks high-precision control~\cite{vqvla}; and rule-based adaptive methods (e.g., FAST~\cite{pertsch2025fast}) introduce decoding instability due to their non-deterministic sequence lengths~\cite{stablefast}.

To address these issues, we introduce FACT (Flow-matching Action Tokenizer). 
Our key insight is to combine VQ-VAE-style discretization with a flow-matching decoding head~\cite{liu2022flow}. 
By shifting the burden of fine-grained motion generation from the discrete latent resolution to a generative decoding process, FACT enables the VLM to plan in a compact, stable discrete space while recovering high-fidelity, continuous trajectories via Ordinary Differential Equation integration.

Based on FACT, we propose the GenieReasoner system, which is illustrated in \Cref{fig:framework}. 
The system operates in two phases:(1) Training (\Cref{fig:framework}a): We employ a unified pipeline that jointly trains the VLM on multimodal reasoning and robotic control. 
By tokenizing continuous action sequences into a sequence of discrete action tokens, the VLM backbone learns to map multimodal observations and high-level instructions to compact action tokens.
(2) Inference (\Cref{fig:framework}b): During deployment, the discrete action codes generated by the VLM backbone are mapped back to continuous control signals via the FACT decoder.
By solving the probability flow ODE, the decoder ensures high-precision manipulation that is semantically grounded in the task instruction.
This integrated framework allows GenieReasoner to leverage the reasoning prowess of VLMs while maintaining the dexterity and precision required for robotic tasks.

\subsection{FACT: \textbf{F}low-matching \textbf{Ac}tion \textbf{T}okenizer}
\label{sec:fact}
\begin{figure}[t]
    \centering
    \includegraphics[width=0.9\linewidth]{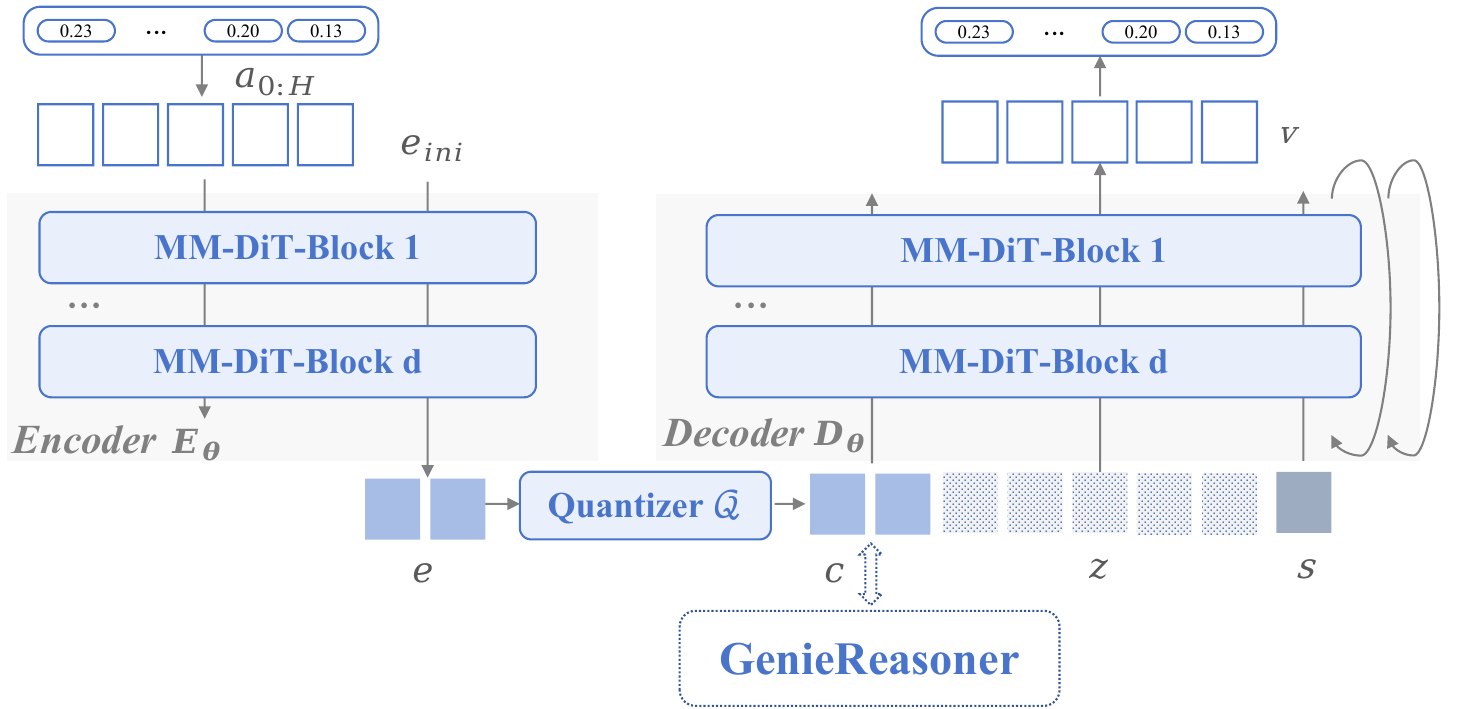}
    \caption{The FACT Action Tokenizer. We discretize continuous robot actions into compact tokens via a VQ-encoder, enabling autoregressive modeling by the VLM. To preserve control precision, the decoder utilizes flow matching to reconstruct smooth, continuous trajectories from the quantized tokens and Gaussian noise z.}
    \vspace{-0.25in}
\label{fig:tokenizer}
\end{figure}
The FACT architecture comprises a VQ-encoder \(\mathcal{E}_\theta\) and a flow-matching decoder \(\mathcal{D}_\theta\). 
For the backbone of both components, we employ the Multimodal Diffusion Transformer (MM-DiT) architecture~\cite{peebles2023scalable}, configured to process sequences with a patch size of one.

\paragraph{Encoder and Quantization}
The tokenization process begins with the action chunk input \(\mathbf{a}_{\tau:\tau+H} \in \mathbb{R}^{H \times S}\). The encoder \(\mathcal{E}_\theta\) maps this input to a continuous latent representation \(e \in \mathbb{R}^{L \times D}\), performing both temporal (\(L \leq H\)) and spatial (\(D \leq S\)) compression. As shown in \Cref{fig:tokenizer}, we use zero-initialized queries \(e_{ini}\) to interact with action features $e = \mathcal{E}_\theta(\mathbf{a}_{\tau:\tau+H}, e_{ini})$.
To obtain a discrete representation, we employ a bitwise quantizer following the lookup-free quantization method~\cite{magvit_v2}. The continuous embedding \(e\) is mapped to a discrete code \(c \in \{-1, +1\}^{L \times D}\) based on the sign of its elements $c = \mathcal{Q}(e) = \operatorname{sign}(e)$.

\paragraph{Flow-Matching Decoder}
We train the decoder network \(\mathcal{D}_\theta\) using the Rectified Flow objective~\cite{liu2022flow}. The goal is to learn the velocity field along a straight-line trajectory that transports samples from a standard Gaussian distribution to the data distribution. This trajectory is defined by the linear interpolation:
\begin{equation}
    a^{(t)} = (1-t)z + t a, \quad t \in [0, 1],
\end{equation}
where \(z \sim \mathcal{N}(0, I)\) is the noise sample and \(a\) is the target action chunk. The target constant velocity along this path is simply \(v^{(t)} = \frac{d}{dt}a^{(t)} = a - z\). The decoder is trained to predict this velocity \(v^{(t)}\) given the noisy state \(a^{(t)}\), the flow time step \(t\), and the condition discrete code\(c\):
\begin{equation}
    v^{(t)} = \mathcal{D}_\theta(a^{(t)}, c, t).
\end{equation}
Architecturally, \(t\) is injected into the MM-DiT blocks via AdaLN modulation, allowing the model to adapt its predictions across the integration trajectory.

\paragraph{Training Loss}
\label{para:training_loss}
The training objective combines quantization regularization with the flow-matching loss.
Following \cite{magvit_v2}, we apply entropy and commitment losses to the quantization process. The entropy loss maximizes codebook usage over the empirical distribution of codes within a batch, denoted as \(\bar{p}(c)\):
\begin{equation}
\mathcal{L}_{\text{entropy}} = \mathbb{E}_{a \sim \mathcal{D}}\left[H\bigl(\bar{p}(c \mid a)\bigr)\right] - H\left(\bar{p}(c)\right).
\end{equation}

The commitment loss ensures the continuous embedding stays close to its quantized value:
\begin{equation}
\mathcal{L}_{\text{commit}} = \mathbb{E}_{a \sim \mathcal{D}} \bigg[ \big\|e - \bar{p}(c) \big\|_2^2 \bigg].
\end{equation}
The flow-matching loss for the decoder network \(\mathcal{D}_\theta\) is formulated as a simple mean-squared error objective:
\begin{equation}
\mathcal{L}_{\text{flow}} = \mathbb{E}_{a, z, t \sim \mathcal{U}[0,1]}\left[ \left\| (a - z) - \mathcal{D}_\theta(a^{(t)}, c, t) \right\|_2^2 \right].
\end{equation}

\paragraph{Inference and Integration}
At inference time, the VLA policy $\pi_\theta$ autoregressively predicts the sequence of discrete action codes $\hat{c}$. 
These codes are passed to the pre-trained decoder $\mathcal{D}_\theta$ to generate the continuous action trajectory. 
Specifically, starting from Gaussian noise $\hat{a}^{(t=0)} \sim \mathcal{N}(0, I)$, we numerically integrate the velocity predicted by the decoder:
\begin{equation}
    \hat{a}^{(t+\Delta t)} = \hat{a}^{(t)} + \Delta t \cdot \mathcal{D}_\theta(\hat{a}^{(t)}, \hat{c}, t).
\end{equation}
The reconstructed action $\hat{a}^{(t=1)}$ is subsequently executed by the robot controller. 
This inference procedure bridges high-level reasoning and low-level motor control: the policy generates discrete and semantically grounded action tokens, while the diffusion decoder ensures high-fidelity and physically feasible motion reconstruction.

\section{Experiments}
\label{sec:experiments}
We empirically validate GenieReasoner by addressing four primary research objectives. First, we assess whether the ERIQ metric effectively decouples reasoning from control and serves as a reliable predictor of downstream task success. 
Second, we examine whether our unified approach achieves superior generalizability in real-world environments compared to leading continuous (e.g., $\pi_{0.5}$) and discrete (e.g., $\pi_0$-FAST) baselines.
Third, we verify the capacity of the FACT tokenizer to reconstruct high-fidelity continuous trajectories from compact discrete representations. 
Finally, we analyze the influence of the training curriculum, particularly the stage of embodied data integration, on the final model performance. 
These inquiries are systematically addressed through benchmark evaluations, real-robot experiments, and extensive ablation studies.


\subsection{Implementation Details}
\label{sec:implementation}
We employ a unified training framework to equip the model with both embodied reasoning capabilities and precise control.
We curated a mixed-source dataset combining generic multimodal corpora (e.g., Cambrian-10M~\cite{tong2024cambrian}), embodied reasoning datasets (e.g., NVIDIA Cosmos-Reason~\cite{azzolini2025cosmos}), and large-scale robotic demonstrations. 
Our training pipeline consists of three stages: 
(1) Tokenizer Training: Learning the FACT tokenizer to compress continuous actions; 
(2) Pre-training: End-to-end optimization on a mixture of general VQA, embodied VQA, and tokenized action data; and 
(3) Post-training: Task-specific fine-tuning with co-training objectives to prevent catastrophic forgetting.
Details about the training data and recipes are provided in \Cref{app:implementation_details}.

\subsection{Main Results}
\begin{table}[t!] 
\centering
\footnotesize
\caption{Performance evaluation on the ERIQ Benchmark with \textit{Spatial Perception \& Grounding} (Spatial), \textit{Planning \& Monitoring} (Planning), \textit{Human Intent Understanding} (Human Intent), and \textit{Error Detection \& Recovery} (Error Loop). Best results are in bold.}
\label{tab:eriq_domains}

\setlength{\tabcolsep}{3.5pt} 

\begin{adjustbox}{max width=\linewidth}
\begin{tabular}{lccccc}
\toprule
\textbf{Model} & \textbf{Spatial} & \textbf{Planning} & \textbf{Human Intent} & \textbf{Error Loop} & \textbf{Avg} \\
\midrule

Qwen2.5-VL-3B   & 62.83 & 54.00 & 82.52 & 47.21 & 58.64 \\
Qwen2.5-VL-7B   & 75.18 & 61.35 & 75.66 & 57.66 & 66.69 \\
Qwen3-VL-8B     & 85.56 & 69.46 & 83.19 & 65.04 & 75.53 \\
InternVL-3.5-8B & 76.25 & 60.59 & 82.08 & 58.38 & 66.72 \\
\midrule
RoboBrain2.0-7B & 76.54 & 60.89 & 84.07 & 57.48 & 67.38 \\
Cosmos-Reason1-7B & 74.48 & 61.64 & 82.97 & 63.24 & 67.99 \\
\midrule
Claude-Sonnet-4 & 72.56 & 59.43 & 88.05 & 55.86 & 65.66 \\
GPT-4o-mini     & \textbf{87.98} & 71.22 & 84.74 & 69.37 & 77.61 \\
Gemini-2.5-pro  & 86.96 & 75.29 & 89.38 & 75.86 & 80.55 \\
\midrule
\rowcolor{gray!10} \textbf{Ours-3B} & 82.21 & \textbf{81.33} & \textbf{89.82} & \textbf{81.08} & \textbf{82.72} \\
\bottomrule
\end{tabular}
\end{adjustbox}
\vspace{-0.3cm}
\end{table}
We evaluate GenieReasoner across the two dimensions of abstract \textit{embodied reasoning} and concrete \textit{end-to-end real-world performance}. 
First, we isolate \textit{embodied reasoning capabilities} using a model variant trained exclusively on VQA data. 
This approach allows us to quantify semantic understanding on the ERIQ benchmark without confounding errors from motor execution. 
Second, we assess \textit{end-to-end real-world performance} by evaluating the complete GenieReasoner system trained via the unified protocol in \Cref{sec:implementation}. 
\subsubsection{Embodied Reasoning Capabilities}
We benchmark GenieReasoner against a comprehensive suite of models, categorized into: (1) general open-source VLMs, (2) specialized open-source embodied VLMs, and (3) large-scale commercial models. 
As shown in Table~\ref{tab:eriq_domains}, our model achieves a state-of-the-art average score of 82.72\% on the ERIQ benchmark, significantly outperforming its base model (58.64\%) and surpassing larger commercial models like Gemini-2.5-pro (80.55\%). 
Crucially, GenieReasoner demonstrates balanced proficiency across all four reasoning dimensions, with a particularly dramatic improvement in \textit{Error Detection \& Recovery} (81.08\% vs. 47.21\% for the base model). 
These capabilities underscore our model's capacity for complex real-world task execution. For a granular performance breakdown across the 15 specific sub-tasks, along with results on open-source spatial benchmarks and qualitative assessments, please refer to Appendix~\ref{app:embodied}.

\begin{figure}[h!]
    \centering
    \includegraphics[width=1.0\linewidth]{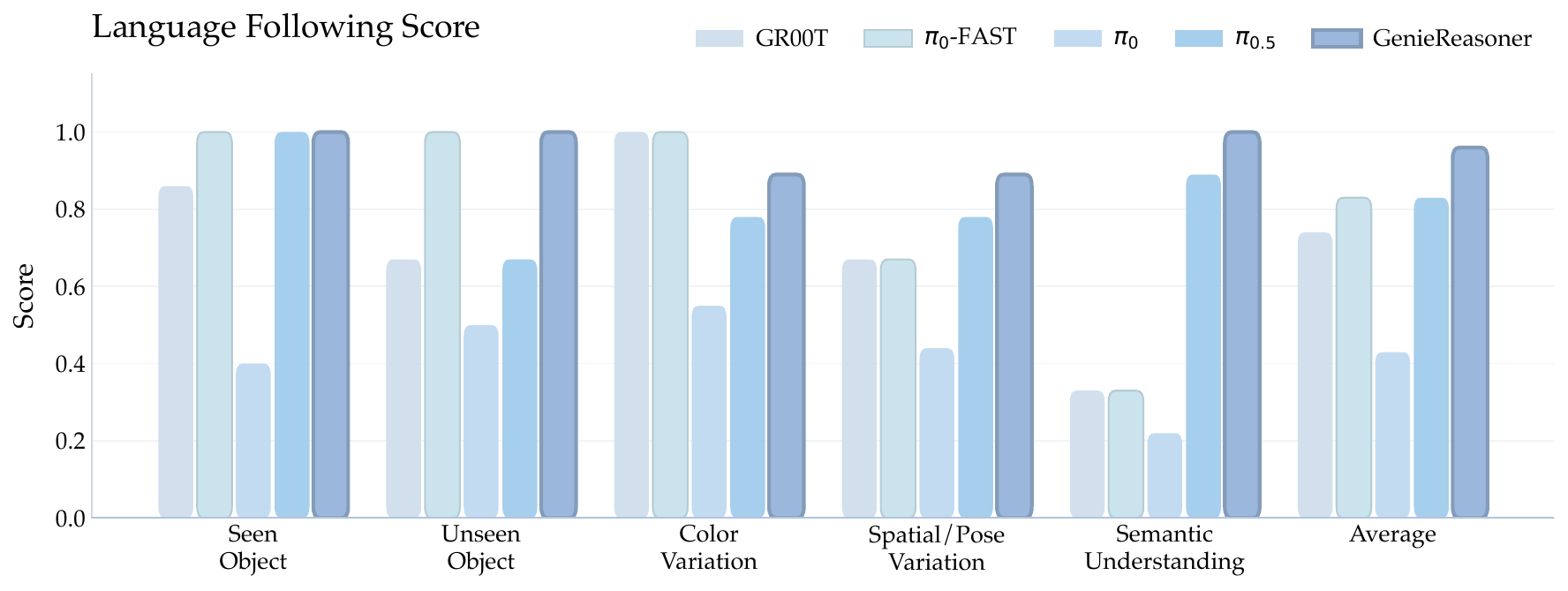}
    \caption{Real-robot language following evaluation which measures the robot's ability to correctly interpret instructions and approach the target object's vicinity.
    }
    \label{fig:realrobot_results_language}
    \vspace{-7pt}
\end{figure}
\begin{figure}[h!]
    \centering
    \includegraphics[width=1.0\linewidth]{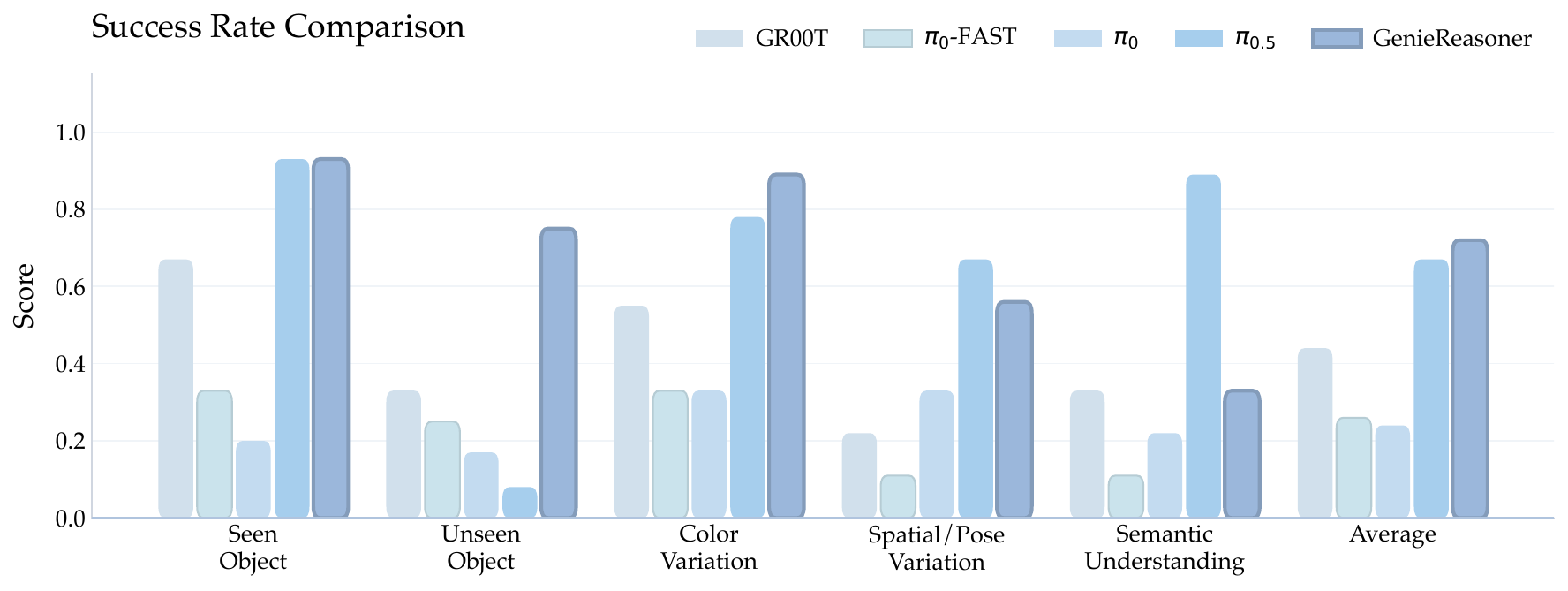}
    \caption{Real-robot success rates which measure the full completion of the task, specifically whether the robot successfully \textit{grasps and manipulates} the target object. 
    }
    \label{fig:realrobot_results_success}
    \vspace{-0.25cm}
\end{figure}

\begin{table*}[t!]
\centering
\caption{Ablation study on training recipes. We evaluate the impact of Action Alignment and Embodied VQA across different training stages. To isolate performance bottlenecks, we report (i) ERIQ scores to measure reasoning capabilities immediately following pre-training, (ii) Language Following to assess semantic grounding, and (iii) Success Rate for end-to-end task completion. Gen. VQA and Emb. VQA denote general and embodied Vision Question Answering datasets. Target, Spatial, and Color represent instruction categories based on direct reference, absolute pose, and visual attributes, respectively.}
\label{tab:e2e_ablation}
\addtolength{\tabcolsep}{-1.7pt} 
\setlength{\aboverulesep}{0pt}
\setlength{\belowrulesep}{0pt}
\renewcommand{\arraystretch}{1.30}
\begin{tabular}{c ccc c| cc ccc ccc}
\toprule

\multirow{2}{*}{\textbf{Exp\#}} & \multicolumn{3}{c}{\textbf{Pre-train}} & \multirow{2}{*}{\textbf{ERIQ} $\uparrow$} & \multicolumn{2}{c}{\textbf{Post-train}} & \multicolumn{3}{c}{\textbf{Language Following} $\uparrow$} & \multicolumn{3}{c}{\textbf{Success Rate} $\uparrow$} \\
\cmidrule(lr){2-4} \cmidrule(lr){6-7} \cmidrule(lr){8-10} \cmidrule(lr){11-13}
 & Gen. VQA & Emb. VQA & Action & \multicolumn{1}{c|}{} & Emb. VQA & Action & Target & Spatial & Color & Target & Spatial & Color \\
\midrule
\#0 &  &  &  & 58.64 &  & \checkmark & 0.24 & 0.05 & 0.17 & 0.05 & 0.00 & 0.06 \\
\#1 & \checkmark & \checkmark &  & \textbf{82.72} &  & \checkmark & 0.28 & 0.15 & 0.25 & 0.04 & 0.00 & 0.03 \\
\midrule
\#2 &  &  & \checkmark & 0.00 &  & \checkmark & 0.57 & 0.45 & 0.52 & 0.12 & 0.20 & 0.07 \\
\#3 & \checkmark & \checkmark & \checkmark & 80.39 &  & \checkmark & 0.78 & \textbf{0.68} & 0.73 & 0.18 & 0.05 & 0.14 \\
\rowcolor{blue!5} 
\#4 & \checkmark & \checkmark & \checkmark & 80.39 & \checkmark & \checkmark & \textbf{0.79} & 0.54 & \textbf{0.91} & \textbf{0.25} & \textbf{0.35} & \textbf{0.22} \\
\bottomrule
\end{tabular}
\vspace{-0.3cm}
\end{table*}

\subsubsection{End-to-End Real World Performance}
\label{subsubsec:real-world}
We validated GenieReasoner on the AgiBot G1 platform against state-of-the-art continuous action baselines ($\pi_0$, $\pi_{0.5}$, GR00T) and the discrete action baseline $\pi_0$-FAST. 
To ensure a rigorous assessment, our evaluation protocol encompasses five settings of increasing complexity. These range from basic \textit{Seen Object} and \textit{Color Variation} tasks to challenging \textit{Unseen Object} and \textit{Spatial/Pose Variation} scenarios involving novel items and extreme orientations. 
Additionally, we included a \textit{Semantic Understanding} setting that requires the model to map abstract instructions, such as ``pick up something to clean the table,'' to specific physical targets. The quantitative results for these experiments are detailed in \Cref{fig:realrobot_results_language,fig:realrobot_results_success}.

Our analysis reveals a critical dichotomy in traditional action representations. The discrete action baseline, $\pi_0$-FAST, achieves superior instruction adherence and high Language Following Scores (\Cref{fig:realrobot_results_language}) by effectively leveraging VLM semantic reasoning to identify targets. However, this semantic proficiency does not guarantee task completion. As illustrated in \Cref{fig:realrobot_results_success}, the success rate for discrete models suffers significantly from quantization artifacts. Such models struggle to represent the continuous precision required for millimeter-level grasping, which leads to frequent alignment failures during physical contact.
In contrast, continuous action baselines like $\pi_{0.5}$ and GR00T exhibit robust motor control once a target is correctly identified but remain highly prone to semantic errors. In complex scenarios involving unseen objects or color shifts, these models frequently approach the wrong target because their action heads lack direct access to the rich reasoning representations of the VLM. This limitation results in a weak connection between perception and control. GenieReasoner resolves this tension by aligning the action space with the VLM's discrete reasoning while utilizing a flow-matching decoder for high-fidelity reconstruction. This unified design matches the manipulation success rates of continuous policies while maintaining or surpassing the instruction following accuracy of discrete architectures. 
For qualitative visualizations of more real-world deployments with different embodiedments, please refer to the Appendix~\ref{app:real_world}.

\subsection{Ablation Study}
To systematically validate the contributions of core design components, we conduct ablation studies targeting the action tokenizer and the training recipes. Specifically, we aim to isolate the impact of the FACT tokenizer on reconstruction fidelity and to quantify the necessity of aligning embodied reasoning data with action supervision.
To facilitate rigorous variable isolation and assess generalization within a reproducible framework, we conduct our training recipe ablations using the high-fidelity \texttt{GenieSim} environment~\cite{2025geniesim}. This approach allows for a systematic evaluation of data composition strategies while mitigating the confounding noise inherent in real-world robotic deployments.

\subsubsection{Ablation on FACT Tokenizer}
\begin{figure}[h!]
    \centering
    \includegraphics[width=0.7\linewidth]{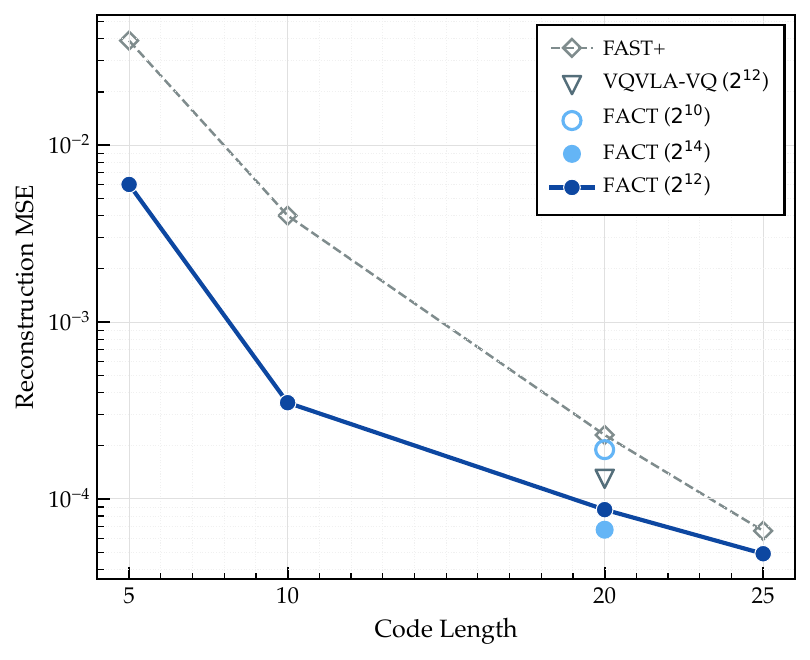}
    \caption{Comparison of Reconstruction Mean Squared Error. Values in parentheses (e.g., $2^{12}$) indicate the vocabulary size used for each configuration.
    }
    \label{fig:exp_UniAcT}
    \vspace{-0.3cm}
\end{figure}

We evaluate FACT's reconstruction fidelity against the FAST+ baseline across varying code lengths (Figure.~\ref{fig:exp_UniAcT}). While the error of FAST+ spikes precipitously as token counts decrease, FACT maintains superior stability, achieving an order-of-magnitude lower error in compact sequences. This resilience confirms that FACT provides a more expressive mapping than BPE-based compression. To isolate the impact of our training objective, we compare FACT against a VQVAE-based tokenizer (using standard MSE loss, same as VQ-VLA~\cite{vqvla}) with an identical architecture. FACT ($2^{12}$) outperforms this MSE-based variant, validating that FACT is more effective for capturing intricate action trajectories.

\subsubsection{Ablation of different training recipes}
We evaluate the training recipe on a similar Open-Set Pick task, mirroring the real-world experiment in \Cref{subsubsec:real-world} with \textit{Target}, \textit{Color}, and \textit{Spatial} instructions.
Similarly, to isolate the source of improvements, we decouple performance into \textit{Language Following} and \textit{Task Success}, post-training all variants on 1,200 real-world episodes.

\paragraph{Embodied VQA as a Foundation}
Comparing the baseline (Exp \#0) with Embodied VQA pre-training (Exp \#1) reveals that VQA data is a prerequisite for grounding. While both models yield negligible success rates ($<5\%$) due to the lack of action alignment, Exp \#1 achieves a dramatically higher ERIQ score (82.72 vs. 58.64). 
This indicates that while Embodied VQA data is insufficient for motor execution in isolation, it establishes the necessary spatial and semantic foundation required for instruction adherence.

\paragraph{Action Alignment and the ERIQ Advantage}
Introducing Action Alignment (Exp \#2, \#3) bridges the gap to physical execution, triggering a performance leap (e.g., 20\% success on spatial tasks). Crucially, these results validate ERIQ as a diagnostic tool: the high reasoning score in Exp \#1 accurately foreshadowed the superior end-to-end performance realized in the fully trained Exp \#4.
This confirms that ERIQ serves as a reliable and efficient predictor of downstream success, enabling the evaluation of reasoning capabilities without the high computational overhead of full action training.

\paragraph{Optimal Synthesis and Post-training}
Finally, Exp \#4 achieves the highest overall performance, surpassing Exp \#3 in both success rates and instruction following. The decisive factor is the inclusion of Embodied VQA data during post-training. Unlike the ``Action Only'' approach in Exp \#3, this joint optimization prevents the catastrophic forgetting of reasoning capabilities, ensuring that high-level semantic understanding is preserved to effectively guide precise physical execution.

\section{Conclusion}
In this work, we demonstrated the necessity of evaluating VLMs independently from action policies and introduced the ERIQ, a large-scale benchmark designed to systematically diagnose embodied reasoning capabilities across four critical dimensions. 
To bridge the gap between semantic reasoning and execution, we proposed GenieReasoner, a unified VLA architecture powered by FACT. By converting continuous control into a high-fidelity discrete representation, this design resolves the reasoning-precision trade-off, enabling the consistent co-optimization of high-level planning and low-level action within a single autoregressive framework.


\bibliographystyle{plainnat}
\bibliography{ref}
\newpage
\clearpage
\appendix
\section{appendix}

\begin{table*}[t]
\centering
\caption{Comparison of ERIQ with other embodied reasoning benchmarks. Data is strictly aligned with the source: ERIQ (Ours) achieves full support ($\CIRCLE$) across all reasoning dimensions and physical data attributes compared to existing benchmarks. Reg. denotes regression metrics for continuous values.}
\label{tab:benchmark_comparison}
\renewcommand{\arraystretch}{1.1}
\resizebox{\textwidth}{!}{%
\begin{tabular}{lccccccccc}
\toprule
\rowcolor{gray!5} & & \textbf{Test Scale} & \multicolumn{2}{c}{\textbf{Data Source}} & \multicolumn{4}{c}{\textbf{Reasoning Dimensions}} & \\ \cmidrule(lr){4-5} \cmidrule(lr){6-9}
\rowcolor{gray!5} \multirow{-2}{*}{\textbf{Benchmark}} & \multirow{-2}{*}{\textbf{Type}} & \textbf{(QA Pairs)} & \textbf{Robo View} & \textbf{Real-world Data} & \textbf{Spatial} & \textbf{Plan} & \textbf{Error} & \textbf{Human Intent} & \multirow{-2}{*}{\textbf{Evaluation Method}} \\ \midrule

RoboSpatial-home~\cite{song2025robospatial} & Image & 350 & \Circle & \CIRCLE & \CIRCLE & \Circle & \Circle & \Circle & MC/Rule \\
RoboSpatial-val~\cite{song2025robospatial} & Image & 6,000 & \Circle & \CIRCLE & \CIRCLE & \Circle & \Circle & \Circle & MC/Rule \\

EmbodiedBench~\cite{yang2025embodiedbench} & Image & 1128 & \CIRCLE & \Circle  & \CIRCLE & \CIRCLE &\LEFTcircle & \Circle & Policy S.R. \\

EmbodiedEval~\cite{EmbodiedEval} & Image & 328 & \Circle & \Circle & \CIRCLE & \CIRCLE & \Circle & \Circle & MC \\

EgoThink~\cite{cheng2024egothink} & Image & 700 & \Circle & \CIRCLE & \CIRCLE & \CIRCLE & \Circle & \Circle & LLM based Open-ended \\

EO-bench~\cite{qu2025eo} & Image & 648 & \CIRCLE & \CIRCLE & \CIRCLE & \CIRCLE & \Circle & \Circle & MC \\

MMRo~\cite{li2024mmro} & Image & 26,175 & \CIRCLE & \CIRCLE & \CIRCLE & \CIRCLE & \Circle & \Circle & MC/LLM+Human based Open-ended/Reg. \\ \midrule

Cosmos-Reason1~\cite{azzolini2025cosmos} & Video & 1,205 & \CIRCLE & \CIRCLE & \CIRCLE & \CIRCLE & \LEFTcircle & \Circle & MC \\

EgoPlan~\cite{chen2023egoplan} & Video & 1,584 & \Circle & \CIRCLE & \Circle & \CIRCLE & \Circle & \Circle & MC \\
ERQA~\cite{gemini_robotics} & Video & 400 & \CIRCLE & \CIRCLE & \CIRCLE & \CIRCLE & \Circle & \Circle & MC \\ 
RoboVQA~\cite{sermanet2024robovqa} & Video & 1,000 & \CIRCLE & \CIRCLE & \Circle & \CIRCLE & \LEFTcircle & \Circle & Human based Open-ended \\
OpenEQA~\cite{majumdar2024openeqa} & Video & 2,193 & \Circle & \CIRCLE & \CIRCLE & \CIRCLE & \Circle & \Circle & LLM based Open-ended \\
VidEgoThink~\cite{cheng2024videgothink} & Video & 4,993 & \Circle & \CIRCLE & \CIRCLE & \CIRCLE & \LEFTcircle & \Circle & MC/LLM-based Open-ended/Reg.\\ 
ShareRobot-Bench~\cite{robobrain} & Video & 3,442 & \CIRCLE & \CIRCLE & \CIRCLE & \CIRCLE & \Circle & \Circle & LLM based Open-ended/Reg. \\ 
RoboBench~\cite{luo2025robobench} & Video & 6,092 & \CIRCLE & \CIRCLE & \CIRCLE & \CIRCLE & \LEFTcircle & \LEFTcircle & MC/Rule/LLM based Open-ended/Reg. \\ \midrule

\rowcolor{blue!5}
\textbf{ERIQ (Ours)} & \textbf{Video} & \textbf{6,052} & \CIRCLE & \CIRCLE & \CIRCLE & \CIRCLE & \CIRCLE & \CIRCLE & \textbf{MC} \\ 
\bottomrule
\end{tabular}%
}
\vspace{0.4em} \\
{\footnotesize \CIRCLE: Fully Supported\ \ \LEFTcircle: Partially Supported  \ \ \Circle: Not Supported }
\end{table*}
\subsection{ERIQ Benchmark}
\label{app:eriq}
\begin{figure}[h]
    \centering
    \includegraphics[width=1.\linewidth]{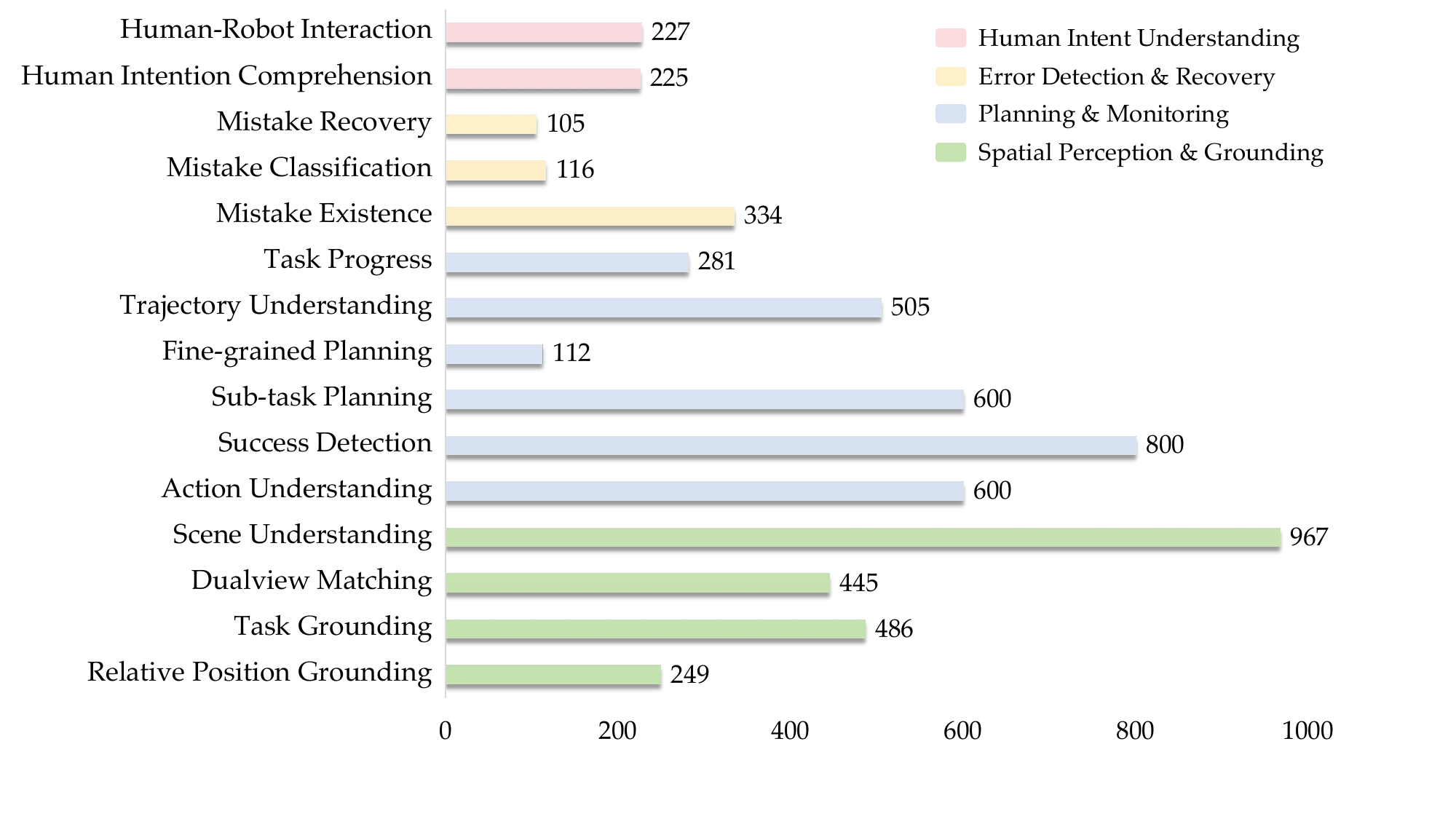}
    \caption{
    Distribution of the ERIQ benchmark across its 15 fine-grained sub-tasks.
    }
    \label{fig:benchmark_distribution}
\end{figure}

\begin{figure*}[ht!]
    \centering
    \includegraphics[width=0.99\linewidth]{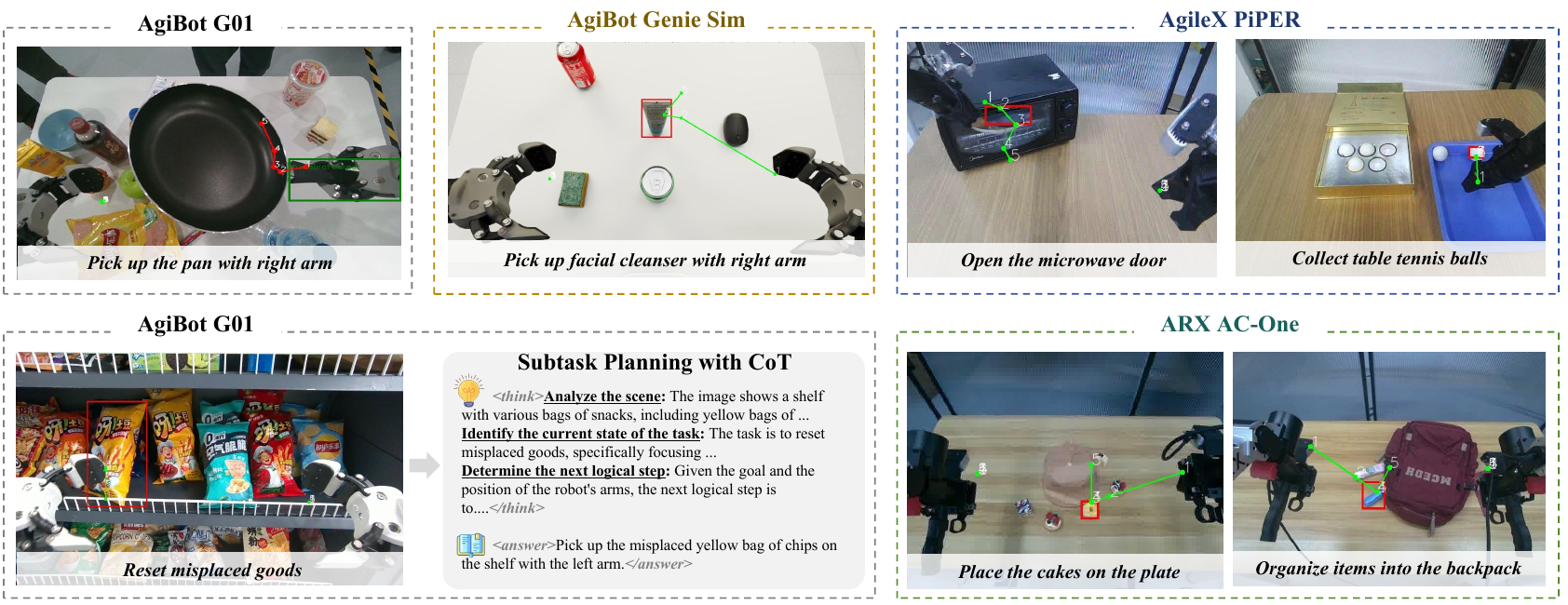}
    \caption{Qualitative visualization of the VLM’s multi-task reasoning capabilities. We display spatial reasoning, subtask planning, and Chain-of-Thought (CoT) inference across diverse embodiments (AgiBot G01, AgiBot Genie Simulation, AgileX, ARX). These results are generated in a zero-shot setting, demonstrating robust spatial and causal reasoning without task-specific adaptation. 
    }
    \label{fig:embodied_cot_viz}
\end{figure*}
\subsubsection{Benchmark Details}
\label{app:eriq_detail}
Each sub-task in ERIQ adheres to a standardized multiple-choice or binary (Yes/No) format, enabling deterministic, straightforward evaluation. 
The specific number of QA pairs for each sub-task is detailed in \Cref{fig:benchmark_distribution}.
The dataset is constructed from over 100 distinct task scenarios spanning five domains: household (35\%), restaurant (20\%), supermarket (20\%), industrial (15\%), and office (10\%). 
To assess multi-modal reasoning comprehensively, the dataset integrates three modalities: Single Image (53\%) for foundational perception, Sequential Images (26\%) for temporal dynamics, and Interleaved Image–Text Sequences (21\%) for context-rich reasoning.

\subsubsection{Benchmark Comparison}
\label{app:eriq_compare}
Despite the expanding scope of existing embodied reasoning benchmarks, they frequently treat task failure as a terminal diagnostic state, overlooking the closed-loop reasoning required for actionable error recovery.
Furthermore, dimensions such as human intent inference and collaborative reasoning remain largely absent from current assessments.
As illustrated in \Cref{tab:benchmark_comparison}, ERIQ establishes a new frontier in embodied reasoning by providing a more comprehensive evaluation suite to date. 
Unlike existing benchmarks that offer fragmented support, ERIQ is the only framework to provide exhaustive, full coverage (\CIRCLE) across all four critical reasoning dimensions: \textit{Spatial Awareness}, \textit{Temporal Planning}, \textit{Error Detection \& Recovery}, and \textit{Human Intent Understanding}.
In stark contrast, even large-scale benchmarks such as MMRo~\cite{li2024mmro} or specialized ones like EgoPlan~\cite{chen2023egoplan} focus primarily on fundamental spatial or planning tasks, leaving high-order cognitive dimensions like Error Recovery and Human Intent Understanding either entirely absent or only partially addressed (\LEFTcircle).
ERIQ comprises 6,052 QA pairs, all of which are derived entirely from authentic, real-world robotic trials to ensure a faithful first-person \textit{Robo View} perspective.
By adopting a deterministic Multiple Choice (MC) evaluation protocol, ERIQ eliminates the subjectivity and noise inherent in LLM-based scoring or human-based open-ended evaluations, providing a rigorous and reproducible diagnostic tool for advancing robust, general-purpose VLA models.

\begin{table*}[t!]
\centering
\footnotesize
\caption{Overall performance evaluation. The table is split into two parts. Part I covers Open-source Spatial Benchmarks and the \textit{Spatial Perception \& Grounding
} subset of  ERIQ Benchmark (including Scene Understanding, Dualview Matching, Task Grounding, and Relative Position Grounding). Part II covers the remaining ER-6K Benchmark subsets: \textit{Planning \& Monitoring
} (Action Understanding, Success Detection, Subtask Planning, Fine-grained Planning, Trajectory Analysis, and Task Progress), \textit{Human Intent Understanding} (Human Intention Comprehension and Human-Robot Interaction), and \textit{Error Detection \& Recovery} (Mistake Existence, Mistake Classification, and Mistake Recovery).}
\label{tab:eriq_detailed}

\setlength{\tabcolsep}{11pt} 
\begin{adjustbox}{width=\textwidth}
\begin{tabular}{l | cccc | cccc }
\toprule
\multirow{2}{*}{\textbf{Model}} & 
\multicolumn{4}{c|}{\textbf{Open-Source Spatial Benchmarks}} & 
\multicolumn{4}{c}{\textbf{ERIQ Benchmark}} \\
\cmidrule(lr){2-5} \cmidrule(lr){6-9}

& CV-Bench & EmbSpa & BLINK-S & BLINK-R & 
Scene Und. & Dualview & Task Grd. & Rel. Pos. \\
\midrule

Qwen2.5-VL-3B & 71.29 & 62.01 & 81.12 & 66.94 & 78.49 & 37.53 & 60.08 & 52.61 \\
Qwen2.5-VL-7B & 80.78 & 70.47 & \textbf{90.21} & 79.03 & 86.66 & 50.56 & 75.93 & 73.09 \\
Qwen3-VL-8B & 85.69 & 78.32 & 86.71 & \textbf{87.10} & \textbf{90.90} & 81.35 & 88.27 & 67.07 \\
InternVL-3.5-8B & 82.16 & 75.11 & 86.01 & 79.84 & 85.94 & 52.58 & 81.69 & 70.28 \\
\midrule
RoboBrain2.0-7B   & \textbf{86.44} & 75.80 & 79.02 & 83.87 & 85.73 & 57.53 & 78.40 & 71.08 \\
Cosmos-Reason1-7B   & 75.20 & 68.90 & 84.62 & 79.84 & 85.94 & 53.48 & 74.69 & 67.07 \\
\midrule
Claude-Sonnet-4   & 75.80 & 64.29 & 79.72 & 72.58 & 79.83 & 64.94 & 66.26 & 70.28 \\
GPT-4o-mini   & 85.21 & 78.29 & 88.11 & 79.03 & 88.00 & \textbf{93.48} & 90.12 & 73.90 \\
Gemini-2.5-pro& 84.59 & \textbf{78.74} & 84.62 & 79.03 & 88.42 & 89.89 & 89.71 & 70.68 \\
\midrule
\rowcolor{gray!10} \textbf{Ours-3B} & 83.48 & 68.52 & 84.62 & 86.29 & 84.18 & 68.54 & \textbf{93.21} & \textbf{77.51} \\
\bottomrule
\end{tabular}
\end{adjustbox}

\vspace{3pt}

\setlength{\tabcolsep}{5.5pt}
\begin{adjustbox}{width=\textwidth}
\begin{tabular}{l | cccccc | cc | ccc | c}
\toprule
\multirow{2}{*}{\textbf{Model}} & 
\multicolumn{11}{c|}{\textbf{ERIQ Benchmark}} & 
\multirow{2}{*}{\textbf{ERIQ-Avg}} \\
\cmidrule(lr){2-12}

& Act. Und. & Success & Subtask & Fine-grained & Traj. & Progress & 
Intention & Interact. & 
Exist. & Classify & Recov. & \\
\midrule

Qwen2.5-VL-3B & 65.50 & 52.75 & 55.67 & 46.43 & 57.62 & 25.98 & 86.22 & 78.85 & 34.43 & 73.28 & 59.05 & 58.64 \\
Qwen2.5-VL-7B & 76.83 & 62.62 & 60.67 & 54.46 & 60.40 & 24.20 & 73.78 & 77.53 & 50.30 & 72.41 & 64.76 & 66.69 \\
Qwen3-VL-8B & 89.17 & 69.38 & 66.33 & 65.18 & 63.76 & 33.45 & 89.78 & 76.65 & 52.99 & 86.21 & 80.00 & 75.53 \\
InternVL-3.5-8B & 78.33 & 55.00 & 59.50 & 59.82 & 58.42 & 27.40 & 80.89 & 83.26 & 50.60 & 74.14 & 65.71 & 66.72 \\
\midrule
RoboBrain2.0-7B   & 82.33 & 56.88 & 61.00 & 51.79 & 58.22 & 24.56 & 84.89 & 83.26 & 50.90 & 76.72 & 57.14 & 67.38 \\
Cosmos-Reason1-7B   & 86.33 & 54.75 & 65.00 & 53.57 & 63.37 & 22.78 & 84.00 & 81.94 & 58.08 & 69.83 & 72.38 & 67.99 \\
\midrule
Claude-Sonnet-4   & 67.67 & 61.50 & 61.33 & 62.50 & 55.84 & 32.03 & 92.89 & 83.26 & 37.13 & 84.48 & 83.81 & 65.66 \\
GPT-4o-mini   & 84.67 & 71.63 & 65.17 & 71.43 & 68.91 & 49.82 & 89.78 & 79.74 & 66.77 & 77.59 & 68.57 & 77.61 \\
Gemini-2.5-pro& 89.83 & 67.37 & 76.67 & \textbf{81.25} & \textbf{75.05} & \textbf{62.28} & 91.11 & \textbf{87.67} & 67.07 & 90.52 & \textbf{87.62} & 80.55 \\
\midrule
\rowcolor{gray!10} \textbf{Ours-3B} & \textbf{96.67} & \textbf{85.25} & \textbf{90.50} & 55.36 & 73.86 & 51.60 & \textbf{96.44} & 83.26 & \textbf{75.45} & \textbf{93.10} & 85.71 & \textbf{82.72} \\
\bottomrule
\end{tabular}
\end{adjustbox}
\label{tab:aer_open_benchmarks}
\end{table*}
\subsection{Detailed Implementation Settings}
\label{app:implementation_details}

\paragraph{Training Data Composition}
To preserve general visual-language capability while enhancing embodied understanding and reasoning, we curated a mixed-source training dataset that combines generic multimodal data with embodied reasoning-specific datasets.
For generic VLM training, we use open-source multimodal datasets including \emph{Cambrian-10M}~\cite{tong2024cambrian}, \emph{LLaVA-OneVision}~\cite{li2024llava}, \emph{Describe Anything}~\cite{lian2025describe}, \emph{CogVLM-SFT-311K}~\cite{wang2024cogvlm}, and \emph{BLIP3-Grounding-50M}~\cite{xue2024xgen}, covering images, videos, dialogue-style QA, and detection-template QA. Parts of these corpora are converted and translated to support bilingual (Chinese/English) training. 
To strengthen embodied understanding, we further incorporate several open-source embodied-intelligence datasets, such as \emph{NVIDIA Cosmos-Reason}~\cite{azzolini2025cosmos}, \emph{ShareRobot}~\cite{robobrain}, \emph{Robo2VLM}~\cite{chen2025robo2vlm}, \emph{EmbSpatial-SFT}~\cite{du2024embspatial}, and \emph{ManipulationVQA-60K}~\cite{chen2025robo2vlm}, which span planning, trajectory reasoning, spatial understanding, and embodied VQA.

While publicly available datasets provide broad embodied knowledge, they often exhibit a distributional mismatch with the specific egocentric perspectives required for real-world deployment. To ensure homologous alignment between reasoning and the robot's observation space, we curated an embodied reasoning dataset based on \emph{AgiBot World}~\cite{bu2025agibot}, which includes (i) 2D trajectory data, where action sequences are projected onto images from three views (top-head and dual wrists); (ii) grounding annotations combining human labels with automated labels for key-object recognition and localization; (iii) sub-task planning data, which are fully annotated and linguistically rewritten to ensure diversity; and (iv) scene understanding, alongside other essential auxiliary tasks. 

For action pre-training, we utilize large-scale embodied demonstrations from the  \emph{AgiBot World} platform, multi-embodiment data collected on ARX and AgileX robots, and various open-source manipulation datasets, providing substantial diversity across embodiments, scenes, and actuation styles.

\paragraph{Detailed Training Recipe}
We implement a three-stage training strategy comprising two pre-training stages followed by a final post-training stage.
\begin{itemize}
    \item {Stage 1 (Tokenizer Training):} We train the FACT tokenizer as in \Cref{fig:tokenizer} with the loss function described in \Cref{sec:fact}.
    \item {Stage 2 (Joint Pre-training):} We perform end-to-end joint training on a mixture of general VQA data, embodied VQA data, and tokenized action data. This joint pre-training enables the model to simultaneously acquire spatio-temporal grounding and motor control capabilities within a unified representation space.
    \item {Stage 3 (Post-training):} In the final stage, we conduct task-specific post-training while maintaining the co-training objective—mixing embodied VQA and action data—to stabilize alignment and prevent the catastrophic forgetting of reasoning capabilities.
\end{itemize}
\begin{figure*}[h!]
    \centering
    \includegraphics[width=1.\linewidth]{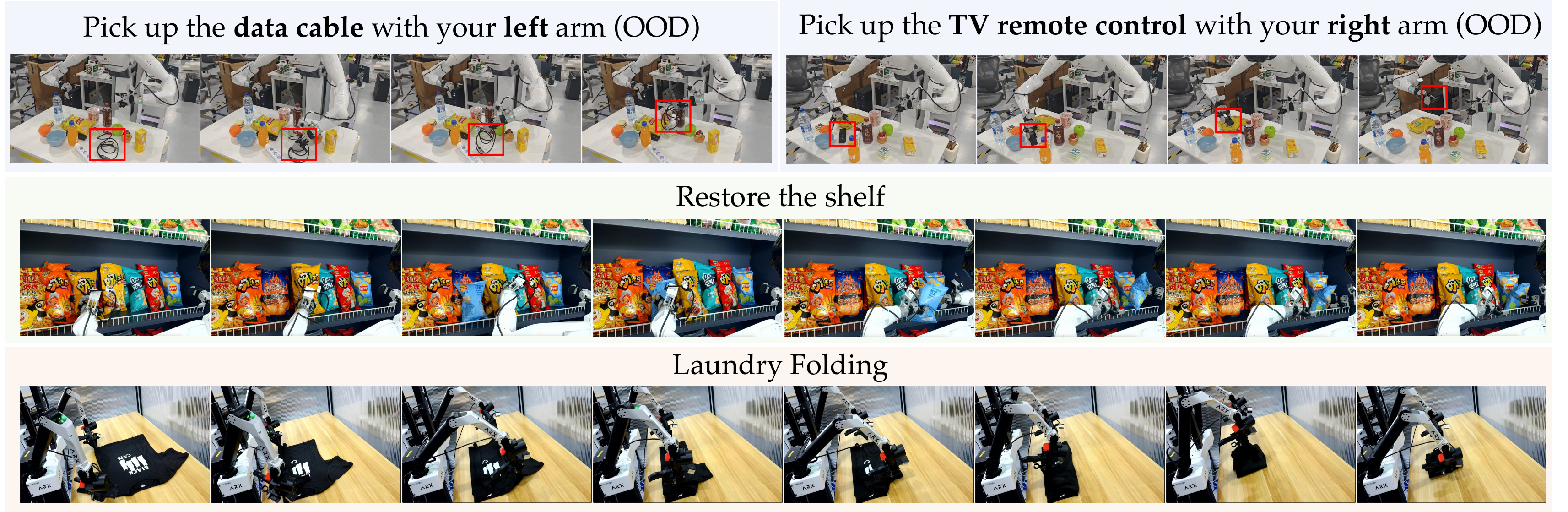}
    \caption{Visualization of different manipulation tasks in real world.}
    \label{fig:realrobot_vis}
    \vspace{-10pt}
\end{figure*}
\subsection{Embodied Reasoning Experiments}
\label{app:embodied}

\subsubsection{Detailed Embodied Reasoning Capabilities Evaluation}

Table~\ref{tab:eriq_detailed} provides a comprehensive breakdown of performance, split into two parts: Part I covers open-source spatial benchmarks and the Spatial ERIQ subset, while Part II details the remaining ERIQ reasoning dimensions. On established external benchmarks such as CV-Bench and BLINK, our model generally outperforms the base model (Qwen2.5-VL-3B) despite not being explicitly trained on these datasets. For instance, on CV-Bench and BLINK-R, GenieReasoner achieves 83.48\% and 86.29\%, respectively (compared to 71.29\% and 66.94\% for the base model), indicating that our embodied pre-training enhances general spatial reasoning capabilities.
On our specific fine-grained sub-tasks:
\begin{itemize}
    \item \textit{Spatial Perception \& Grounding:} Within our proposed benchmark, the model effectively grounds abstract instructions in physical space, achieving 93.21\% on Task Grounding and 84.18\% on Scene Understanding, proving its robustness in cluttered, real-world environments.
    
    \item \textit{Error Detection \& Recovery:} In this critical closed-loop domain, the model demonstrates robust diagnostic skills, scoring 93.10\% on Mistake Classification and 85.71\% on Mistake Recovery. This represents a significant leap over the base model, enabling actionable fixes for execution failures.
    
    \item \textit{Human Intent Understanding:} The model achieves its highest domain score here (Avg: 89.82\%), driven by a near-perfect 96.44\% on Human Intention Comprehension, significantly surpassing the base Qwen2.5-VL-3B (86.22\%).
    
    \item \textit{Planning \& Monitoring:} We observe strong performance in logical sequencing, with scores of 96.67\% in Action Understanding and 90.50\% in Subtask Planning. Additionally, the model excels at Success Detection (85.25\%), effectively identifying task completion.
\end{itemize}

\subsubsection{Qualitative results of Generalized Embodied Reasoning}
\label{app:qualitative_results}
As shown in \Cref{fig:embodied_cot_viz}, we qualitatively demonstrate the model's robust generalization and spatial reasoning capabilities across diverse robotic platforms. We visualize three core outputs: task-oriented grounding, trajectory prediction, and chain-of-thought (CoT) inference, across a variety of tasks and scenes. These include real-world deployments on AgiBot G01, AgileX, and ARX platforms, and a high-fidelity simulation environment AgiBot GenieSim\cite{2025geniesim}.
The results reported herein were obtained in a \emph{zero-shot} manner, necessitating no task-specific adaptation. This underscores the robust inherent generalization of our VLM across both spatial and causal reasoning dimensions. 
These findings validate the framework's efficacy in facilitating cross-embodiment transfer and operational robustness within unstructured real-world environments.

\subsection{Real-World Deployment and Evaluation}
\label{app:real_world}

\subsubsection{Operational Details of Real-World Evaluation}
We evaluated the system across diverse environmental configurations in \Cref{subsubsec:real-world}, with each scene containing 4 to 11 objects to simulate the entropy of unstructured workspaces. To test discriminative precision, we introduced distractors with high chromatic and semantic similarity into the workspace. A typical setup involved placing visually analogous items—such as a blue sponge and a blue toy—in close proximity. This forced the model to ground instructions to the specific target based on nuanced semantic cues rather than relying on simple color-based saliency.

Spatial reasoning was further challenged through trials involving identical objects that could only be disambiguated by their relative physical attributes. For instance, we used water bottles of varying heights, requiring the model to resolve linguistic descriptors like ``higher'' and ``lower'' to generate the appropriate physical trajectory. To test the robustness of the vision-action coupling, we also introduced unconventional object orientations not present in the training set, such as a milk carton placed in a horizontal, tipped-over pose. These scenarios helped determine whether the vision encoder captures the invariant semantic identity of the object or is merely overfitted to the canonical upright poses found in standard datasets.

Finally, we addressed a common egocentric bias observed in vision-based controllers, where models tend to heuristically grasp whatever appears directly beneath the gripper in the head-mounted camera's view. We specifically configured our experimental scenes to include proximal distractors in these "low-hanging" areas. This allowed us to verify that the model maintains fidelity to the actual linguistic intent rather than defaulting to proximity-based visual shortcuts.

Our evaluation also revealed that some models often default to grasping the most accessible object, effectively replicating common trajectories rather than reasoning about the instruction. 
In certain cases, this behavior can artificially inflate performance metrics.
For example, when tasked with retrieving either the ``higher'' or ``lower'' bottle, certain models (such as $\pi_{0.5}$) consistently targeted the lower one regardless of the prompt. 
While this resulted in a partial score due to the occasional alignment with the target, it highlights a failure in semantic grounding, where the model prioritizes high-probability motion patterns over actual linguistic intent.

\subsubsection{More Real World Visualization}
To evaluate the cross-embodiment robustness and operational generalization of GenieReasoner, we conducted extensive real-world experiments leveraging both the AgiBot G1 humanoid platform and the ARX AC-One robotic arm. 
The evaluation protocol consists of three core task categories designed to investigate the integration of high-level semantic grounding with fine-grained motor control: 
(i) \textit{fine-grained OOD manipulation}, involving the retrieval of out-of-distribution objects (e.g., data cables, TV remotes) in high-entropy, cluttered environments, extending the evaluation protocol described in \Cref{subsubsec:real-world}; 
(ii) \textit{long-horizon semantic tasks}, such as structured shelf restoration; and 
(iii) \textit{deformable object manipulation}, exemplified by garment folding conducted on the ARX platform. 
These diverse scenarios require the model to translate abstract multimodal instructions into adaptive, high-precision physical trajectories.
Qualitative visualizations of these results across the two distinct robotic embodiments and complex environments are provided in \Cref{fig:realrobot_vis}.
Comprehensive video demonstrations are provided in the supplementary materials.
\end{document}